\pdfoutput=1
\documentclass[runningheads]{llncs}

\usepackage{eccv}

\usepackage{algpseudocode}
\usepackage{algorithm}
\usepackage{multirow}

\usepackage{url}
\usepackage{makecell}
\usepackage[utf8]{inputenc}
\usepackage[export]{adjustbox}  
\usepackage{float}

\usepackage{eccvabbrv}

\usepackage{graphicx}
\usepackage{booktabs}

\usepackage{hyperref}

\usepackage{orcidlink}
\usepackage[misc]{ifsym}

\begin{document}

\title{Joint Alignment and Distillation for \\ Video Generation via Sample-Guided \\ Distribution Matching} 







\author{
Jiuzhou Lin\inst{1,2}\textsuperscript{$\dagger$}
\and Junlong Wu\inst{1,2}\textsuperscript{$\dagger$}
\and Fei Zuo\inst{2,3}
\and Huan Ouyang\inst{2,4}
\and Dewen Fan\inst{2}
\and Boheng Zhang\inst{2}
\and Huaiqing Wang\inst{2}
\and Jia Sun\inst{2}
\and Fan Yang\inst{2}\textsuperscript{*}
\and Houde Liu\inst{1}\textsuperscript{*}
\and Kehai Chen\inst{3}
\and Min Zhang\inst{3}
\and Tingting Gao\inst{2}
\and Han Li\inst{2}
}

\authorrunning{J.~Lin, J.~Wu et al.}
\titlerunning{Video Generation via Sample-Guided Distribution Matching}

\institute{
Tsinghua University \and Kuaishou Technology
\and
Harbin Institute of Technology (ShenZhen)
\and
Beijing University of Posts and Telecommunications
\\
\textsuperscript{$\dagger$} Equal contribution.
\quad
\textsuperscript{*} Corresponding authors
\\
\textsuperscript{\Letter} \email{\{lin-jz24, wu-jl24\}@mails.tsinghua.edu.cn}
}

\maketitle

\begin{abstract}
    Aligning video generative models to human preferences heavily relies on Reinforcement Learning (RL), which suffers from extensive computational overhead. 
    Existing workflows typically treat RL and distillation as disconnected stages: applying RL before distillation incurs prohibitive computational costs, whereas applying RL after distillation frequently leads to model collapse. 
    To overcome these limitations, we propose a unified, single-stage optimization framework grounded in Distribution Matching (DM). 
    In the standard DM framework, distillation updates the model via a gradient direction that minimizes the gap between the real and fake models, guiding generations toward clarity and high fidelity. 
    Building upon this, we introduce DM-Align, which derives a complementary gradient direction to guide the model toward human-preferred samples. 
    Inspired by DPO and GRPO, our method leverages the distributional gap—formulated from either preference pairs or intra-group exploration—to directly construct this preference-guided gradient. 
    By synergizing these two gradient directions, our approach eliminates the need for multi-step reward evaluation and complex ODE-SDE conversions inherent in traditional RL. Comprehensive experiments across multiple foundational video models demonstrate that this sample-guided framework robustly enhances both distillation quality and preference alignment, consistently outperforming both standalone variants and sequential two-stage pipelines.
\end{abstract}

\section{Introduction}
\label{sec:intro}

Recent advances in visual generative modeling, primarily driven by diffusion models \cite{ho2020denoising, song2020denoising} and flow matching models \cite{lipman2022flow, liu2022flow}, have fundamentally transformed image and video generation. 
Both open-source models \cite{polyak2024movie, kong2024hunyuanvideo, wan2025wan} and proprietary commercial systems \cite{kuaishou2024kling, deepmind2024veo2} can now generate high-quality images or videos. 
This progress has established a research paradigm mirroring that of large language models: leveraging open-source foundational models to develop and validate new algorithms.
In this work, we focus on two key research directions in diffusion models:
1) \textit{Distillation} \cite{yin2024one, yin2024improved, chadebec2025flash}, which reduces generation time while maintaining output quality, enabling few-step generation comparable to the original multi-step base models; 
and 2) \textit{Alignment} \cite{liu2025improving, xue2025dancegrpo, li2025mixgrpo}, which fine-tunes models to align with human preferences or aesthetic judgments, often leveraging Reinforcement Learning (RL) techniques like Direct Preference Optimization (DPO) \cite{rafailov2023direct} and Group Relative Policy Optimization (GRPO) \cite{shao2024deepseekmath}.

\begin{figure*}[t]
  \centering
  \includegraphics[width=\linewidth]{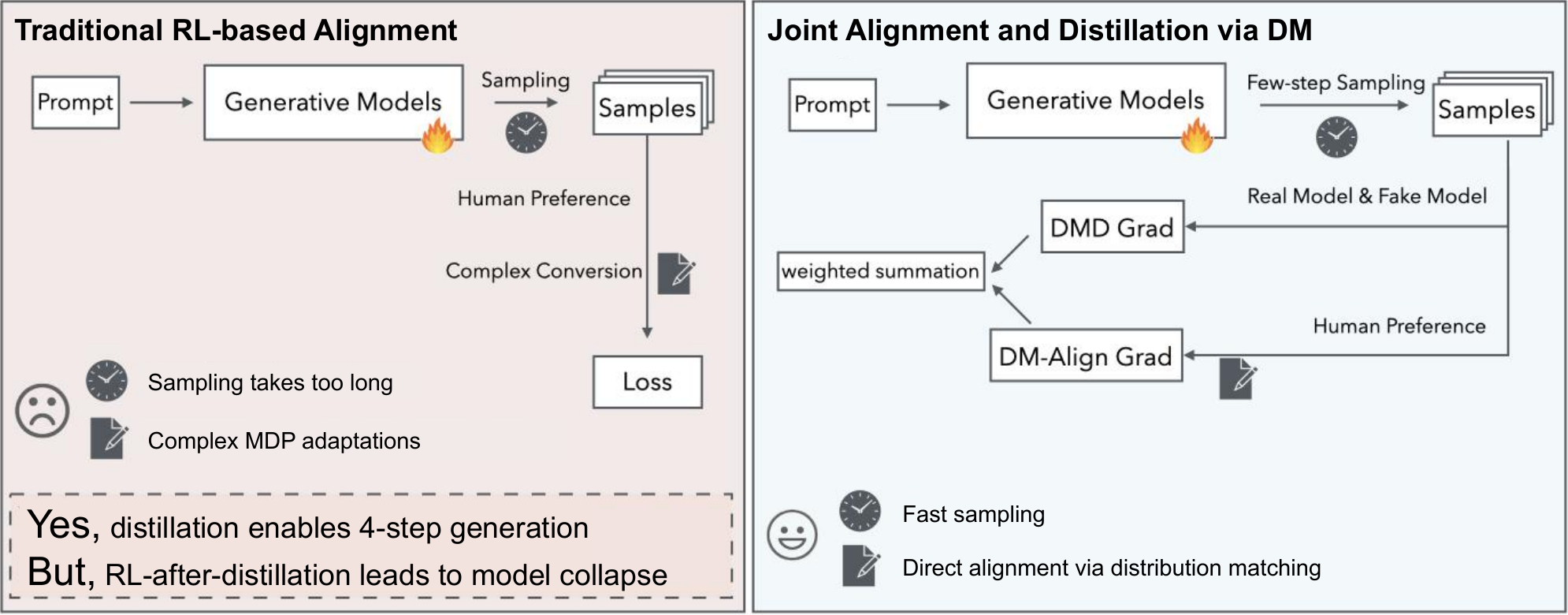}   
    \caption{   
      \textit{Left:} Applying RL before distillation incurs prohibitive multi-step sampling costs and complex MDP adaptations, while applying RL after distillation causes unstable training. 
      \textit{Right:} Our unified Sample-Guided Distribution Matching framework jointly optimizes both objectives in a single stage, enabling fast, high-fidelity preference alignment while entirely bypassing the bottlenecks of traditional RL.
    }
  \label{figs:fig1}
\end{figure*}

As illustrated in Figure~\ref{figs:fig1} (Left), existing workflows typically treat distillation and alignment as disconnected, sequential stages, which inevitably leads to a dilemma. 
On one hand, applying RL \textit{before} distillation \cite{chen2025skyreels} incurs prohibitive computational costs, as obtaining generated samples requires extensive multi-step sampling. 
More fundamentally, existing RL-based methods adapt the diffusion sampling process into the Markov Decision Process (MDP) formulation of RL algorithms. 
This paradigm requires calculating trajectory probabilities along the sampling path, necessitating complex conversions into reverse-time Stochastic Differential Equations (SDEs) \cite{song2020score, albergo2023stochastic} and reliance on Gaussian distribution assumptions. 
On the other hand, applying RL \textit{after} distillation to leverage few-step generation \cite{yin2024one, yin2024improved, luo2025learning} is unstable. 
Fine-tuning a highly compressed distilled model with sparse RL losses frequently destroys the delicate continuous generative mapping, leading to visual artifacts (as illustrated in Figure~\ref{fig:distill_after}).

To overcome these limitations, as shown in Figure~\ref{figs:fig1} (Right), we shift the paradigm from forcing diffusion into an RL framework to unifying both objectives within a joint \textbf{Distribution Matching (DM)} framework. 
From a theoretical perspective, we observe a gradient compatibility between distillation and alignment under the DM objective: both objectives can be expressed as score-space distribution-matching gradients and can therefore be optimized within a shared computational pathway.   
In the standard DM framework, distillation updates the model via a gradient direction that minimizes the gap between the teacher (real) and student (fake) models, guiding generations toward clarity and high fidelity. 
In a parallel manner, alignment can be formulated by measuring the distributional gap between ordinary generated samples and human-preferred samples. 
This yields a complementary gradient direction that naturally shifts the model's overall distribution toward optimal structural preferences.

Motivated by this gradient compatibility, we propose a unified, single-stage optimization framework, introducing \textbf{DM-Align}.  
While DM-Align leverages preference samples inspired by RL (such as win-lose pairs from DPO or intra-group exploration from GRPO), it operates fundamentally as a \textit{sample-guided DM loss}. 
By directly deriving preference-guided gradients based on these sample disparities, our method entirely bypasses the expensive multi-step generation and eliminates the need for complex mathematical adaptations. 
The distillation gradient and the alignment gradient are synergistically aggregated within the same latent space, guiding the generator toward both high fidelity and human preference without the risk of post-distillation collapse.

To summarize, our key contributions are the following:
\begin{itemize}
    \item 
    We identify the gradient compatibility between distillation and preference alignment under the Distribution Matching objective. 
    Based on this, we propose a unified, single-stage framework that optimizes both objectives simultaneously, successfully avoiding the prohibitive computational costs of RL-before-distillation and the catastrophic model collapse inherent in RL-after-distillation.
    
    \item 
    We introduce \textbf{DM-Align}, a sample-guided approach that entirely bypasses the need to adapt the diffusion sampling process into an RL framework. 
    Specifically, we develop \textbf{DM-PairLoss} and \textbf{DM-GroupLoss} as concrete implementations inspired by DPO and GRPO, directly utilizing preference pairs and group rewards to construct efficient gradient guidance.
    
    \item 
    We conduct comprehensive experiments across multiple foundational video models (e.g., Wan 2.1 T2V-1.3B~\cite{wan2025wan} and CogVideoX~\cite{yang2024cogvideox}) and varying reward models (e.g., VideoAlign~\cite{liu2025improving}, HPSv2~\cite{wu2023human}). 
    Our method demonstrates exceptional generalizability and substantially outperforms strong baselines, achieving significant gains in VBench scores (up to +6.55) and human evaluations (up to +48\% net preference) while exhibiting remarkable training efficiency.
\end{itemize}


\section{Related Work}

\subsection{Aligning Diffusion Models and Flow Matching Models}
Motivated by the success of Reinforcement Learning from Human Feedback (RLHF) \cite{ouyang2022training} in large language models, a significant line of work adapts RL algorithms to align visual generation models with human preferences. 
These methods primarily fall into two categories: DPO-based approaches \cite{rafailov2023direct, wallace2024diffusion, yang2311using, liang2024step, zhang2025diffusion, liu2025improving} that rely on annotated positive/negative sample pairs, and PPO/GRPO-based methods \cite{schulman2017proximal, black2023training, fan2023dpok, liu2025flow, xue2025dancegrpo, li2025mixgrpo} that utilize reward models for optimization without paired data.

Although RL algorithms for image generation are theoretically transferable to video tasks, practical implementations often encounter severe training instability and convergence difficulties. 
To address video-specific alignment, methods such as Flow-GRPO \cite{liu2025flow} and DanceGRPO \cite{xue2025dancegrpo} have emerged. 
While achieving competitive results, integrating these models into standard RL frameworks introduces inherent complexities. Specifically, adapting continuous ODE-based generative trajectories into the discrete Markov Decision Process (MDP) formulation requires complex reverse-time SDE conversions and explicit trajectory probability estimations \cite{song2020score}. 
Practically, computing RL losses requires extensive sampling steps (typically 20-40), incurring non-trivial computational overhead. 
Concurrent work MixGRPO \cite{li2025mixgrpo} attempts to mitigate this by enhancing sampling efficiency through SDE-ODE alternation. 
In contrast, we adopt a novel Distribution Matching (DM) perspective, completely circumventing these cumbersome MDP adaptations and multi-step sampling requirements, paving the way to seamlessly unify alignment with highly efficient distillation. More broadly, diffusion-based generative modeling and reward-driven optimization have demonstrated
  effectiveness across a wide range of vision and decision-making tasks ~\cite{li2025diffpcn, yan2025symmcompletion, li2026detailanywhere, gong2026sculpting, wu2025arc, li2025safesim, lin2025keypoint}, which collectively motivate the
  pursuit of more efficient and unified optimization frameworks.

\subsection{Diffusion Distillation}
Accelerating the multi-step inference of visual generative models \cite{ho2020denoising, lipman2022flow} remains a fundamental challenge. 
Existing acceleration techniques can be broadly categorized into advanced samplers \cite{liu2022pseudo, zhao2023unipc, karras2022elucidating}, trajectory distillation \cite{salimans2022progressive, song2023consistency, luo2023latent}, and distribution distillation \cite{yin2024one, yin2024improved, lin2025diffusion}. 
Among these, Distribution Matching Distillation (DMD) \cite{yin2024one, yin2024improved} has emerged as a highly effective paradigm, alternately optimizing real and fake models to minimize the KL divergence between generated and target distributions. 
While initially designed for image synthesis, DMD-based frameworks have been validated by the community as exceptionally stable and effective for video tasks \cite{lightx2v, cheng2025pose}.

Inspired by DMD, recent works \cite{luo2025learning, shao2025magicdistillation, sun2025swiftvideo, gu2025video} have proposed various enhancements primarily to improve distillation efficiency. 
Closely related to our objective are methods like Diff-Instruct++ \cite{luo2024diff} and concurrent works such as FlashDMD \cite{chen2025flashdmdhighfidelityfewstepimage} and DMDR \cite{jiang2025distributionmatchingdistillationmeets}, which also explore the joint optimization of distillation and RL. 
However, these approaches typically perform a loss addition, persistently adapting the diffusion sampling process into standard RL algorithms to compute off-the-shelf RL losses. 
By exploiting the score-space gradient compatibility of distribution matching, we design sample-guided DM losses that formulate preference alignment natively within the DM framework, rather than adding off-the-shelf RL objectives to a distillation loss. Furthermore, unlike these image-centric concurrent efforts, our framework successfully tackles the more complex dynamics of video generation.


\section{Method}
\label{sec3:method}
In Section~\ref{sec3.1:dmd}, we briefly review Distribution Matching Distillation (DMD). 
In Section 3.2, we derive preference alignment as a score-space distribution-matching gradient and show its compatibility with the DMD gradient.
Finally, Section~\ref{sec3.3:unify} details our unified, single-stage framework that simultaneously achieves highly efficient distillation and preference alignment.

\begin{figure*}[t]
  \centering
  \includegraphics[width=\linewidth, page=5]{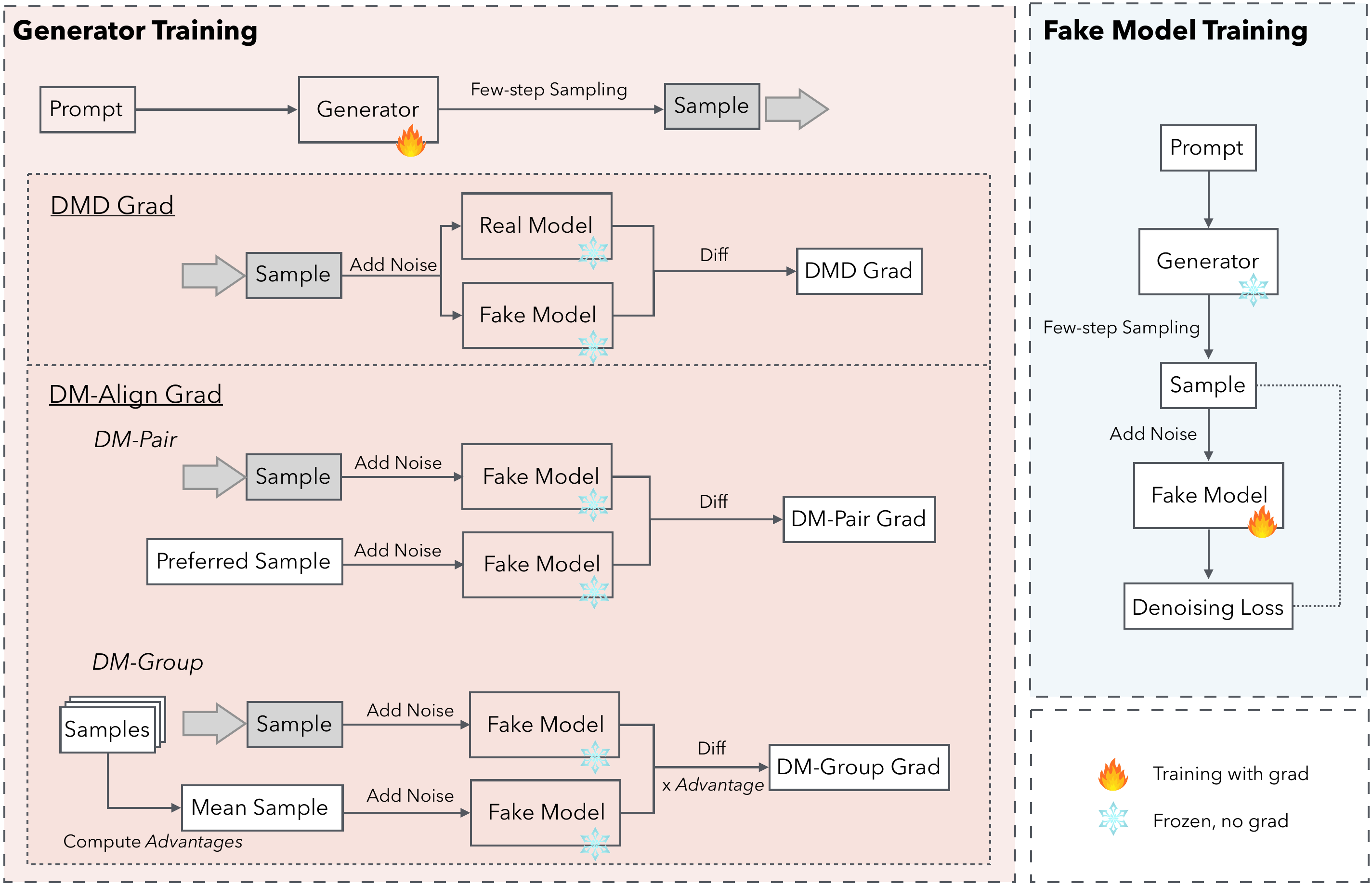}   
  \caption{\textbf{Overview of our unified framework.} \textit{Left:} The generator produces samples via few-step sampling and is updated simultaneously by the distillation gradient and the DM-Align gradient. \textit{Right:} The fake model approximates the generator's current distribution via a standard denoising loss, acting as a robust score estimator for both objectives.}
  \label{figs:fig2}
\end{figure*}

\subsection{Review of Distribution Matching Distillation}
\label{sec3.1:dmd}
Our approach builds upon the foundational Distribution Matching Distillation (DMD) framework \cite{yin2024one, yin2024improved, yin2025slow, huang2025self}. 
The core objective of DMD is to minimize the Kullback-Leibler (KL) divergence between the distilled generator's output distribution ($p_{\mathrm{fake}}$) and the original teacher model's distribution ($p_{\mathrm{real}}$):
\begin{equation}
\mathcal{L}_{\text{DMD}} = D_{\mathrm{KL}}(p_{\mathrm{fake}} \parallel p_{\mathrm{real}}) = \mathbb{E}_{\substack{z \sim \mathcal{N}(0,\mathbf{I}) \\ x = G_{\theta}(z)}} \left[ -( \log p_{\mathrm{real}}(x) - \log p_{\mathrm{fake}}(x) ) \right].
\label{eq:dmd_kl}
\end{equation}
Since directly estimating probability densities is intractable, DMD optimizes the generator parameters $\theta$ by computing the gradient of this divergence. 
This is achieved by perturbing data distributions with Gaussian noise to create overlapping manifolds:
\begin{equation}
\nabla_{\theta} \mathcal{L}_{\text{DMD}} \approx -\mathbb{E}_{t, z} \left[ \Big( \mathbf{s}_{\mathrm{real}}\big(F(G_{\theta}(z), t), t \big) - \mathbf{s}_{\mathrm{fake}}\big(F(G_{\theta}(z), t), t \big) \Big) \frac{dG_{\theta}(z)}{d\theta} \right],
\label{eq:dmd_gradient}
\end{equation}
where $F(\cdot,t)$ denotes the forward diffusion process at noise level $t$, and
$\mathbf{s}_{p}(\mathbf{x}_t,t)=\nabla_{\mathbf{x}_t}\log p_t(\mathbf{x}_t)$ denotes the score function of the noisy distribution $p_t$.
As illustrated in Figure~\ref{figs:fig2}, the real model serves as a fixed teacher, while the fake denoising model $\mu_{\phi}^{\mathrm{fake}}$ is continuously updated to approximate the generator's output distribution and provides the score estimator $\mathbf{s}_{\mathrm{fake}}(\cdot,t)$:
\begin{equation}
\mathcal{L}_{\phi}^{\text{denoise}} = \|\mu_{\phi}^{\text{fake}}(\mathbf{x}_t, t) - \mathbf{x}_0\|_2^2,
\label{eq:denoise_loss}
\end{equation}
where $\mathbf{x}_0 = G_{\theta}(z)$ is the generated sample and $\mathbf{x}_t = F(\mathbf{x}_0, t)$ is its noisy counterpart. 
This framework provides a highly stable mechanism for few-step trajectory compression.

\begin{algorithm*}[t]
\caption{Unified Single-Stage Optimization for Distillation and Alignment}
\label{alg:training}
\begin{algorithmic}[1]
\Require Pretrained model $\mu_{\text{real}}$, conditional input $\mathcal{D} = \{\mathbf{c}\}$, preference data $\mathcal{D}_{\text{pref}}$.
\Require Few-step timesteps $\mathcal{T} = \{\tau_1, \dots, \tau_Q\}$, update ratio $N$, group size $K$.
\State $G_\theta \leftarrow \texttt{copyWeights}(\mu_{\text{real}})$; 
      $\mu_{\text{fake}} \leftarrow \texttt{copyWeights}(\mu_{\text{real}})$ 
\State $iter \leftarrow 1$
\While{training}
    \State Sample batch $\mathbf{z} \sim \mathcal{N}(0, I)^B$ and $\mathbf{c} \sim \mathcal{D}$
    \State Sample distillation timestep $\tau \sim \text{Uniform}(\mathcal{T})$ 
    \State $\mathbf{x}, \mathbf{x}_{\text{final}} \leftarrow \texttt{generateSamples}(G_\theta, \mathbf{z}, \mathbf{c}, \tau)$ \hfill $\triangleright$ Shared generation step
    
    \If{$iter \pmod{N+1} \neq 0$} \hfill $\triangleright$ Phase 1: Update Fake Model ($N$ steps)
        \State Sample denoising timestep $t \sim \mathcal{U}(0, 1)$ and noise $\epsilon \sim \mathcal{N}(0, I)$
        \State $\mathbf{x}_t \leftarrow \alpha_t \cdot \texttt{stopgrad}(\mathbf{x}) + \sigma_t \epsilon$ 
        \State $\mathcal{L}_{\text{denoise}} \leftarrow \texttt{denoisingLoss}(\mu_{\text{fake}}(\mathbf{x}_t, t), \texttt{stopgrad}(\mathbf{x}))$
        \State $\mu_{\text{fake}} \leftarrow \texttt{update}(\mu_{\text{fake}}, \nabla\mathcal{L}_{\text{denoise}})$
        
    \Else \hfill $\triangleright$ Phase 2: Update Generator (1 step)
        \State Sample denoising timestep $t \sim \mathcal{U}(0, 1)$
        \State $\nabla\mathcal{L}_{\text{DMD}} \leftarrow \texttt{DMDGradient}(\mu_{\text{real}}, \mu_{\text{fake}}, \mathbf{x}, t)$ \hfill $\triangleright$ Eq. \ref{eq:dmd_gradient}
        
        \If{using DM-PairLoss}
            \State $\mathbf{x}_+ \leftarrow \texttt{sample}(\mathcal{D}_{\text{pref}}, \mathbf{c})$ 
            \State $\nabla\mathcal{L}_{\text{align}} \leftarrow \texttt{DM-PairGradient}(\mu_{\text{fake}}, \mathbf{x}_{\text{final}}, \mathbf{x}_+, t)$ \hfill $\triangleright$ Eq. \ref{eq:align_gradient}
        \ElsIf{using DM-GroupLoss}
            \State $\mathbf{x}_{\text{group}} \leftarrow \texttt{groupSamples}(\mathbf{x}_{\text{final}}, K)$ 
            \State $\nabla\mathcal{L}_{\text{align}} \leftarrow \texttt{DM-GroupGradient}(\mu_{\text{fake}}, \mathbf{x}_{\text{final}}, \mathbf{x}_{\text{group}}, t)$ \hfill $\triangleright$ Eq. \ref{eq:dm_group_gradient}
        \EndIf
        
        \State $\nabla\mathcal{L}_G \leftarrow \lambda_{\text{DMD}}\nabla\mathcal{L}_{\text{DMD}} + \lambda_{\text{align}}\nabla\mathcal{L}_{\text{align}}$ \hfill $\triangleright$ Synergistic gradient aggregation
        \State $G_\theta \leftarrow \texttt{update}(G_\theta, \nabla\mathcal{L}_G)$
    \EndIf
    
    \State $iter \leftarrow iter + 1$
\EndWhile
\end{algorithmic}
\end{algorithm*}

\subsection{Alignment via Sample-Guided Distribution Matching}
\label{sec3.2:dm_align}
While DMD successfully minimizes the gap to the teacher model, aligning the model with human preferences requires a fundamental paradigm shift. 
Traditional RL approaches attempt to solve this by adapting diffusion sampling into a Markov Decision Process (MDP), necessitating complex trajectory probability estimations and extensive multi-step sampling. 
In contrast, we reveal a score-space gradient compatibility: preference alignment can be natively formulated within the DM framework by matching the generator distribution to a preference-optimal target distribution.

\textbf{Theoretical Formulation.} Following the standard RLHF framework under the Bradley-Terry model, the optimal policy $p^*$ that maximizes expected reward while remaining anchored to a reference distribution $p_{\mathrm{ref}}$ has the closed-form solution \cite{peters2010relative}:
\begin{equation}
    p^*(x) = \frac{1}{Z} p_{\mathrm{ref}}(x) \exp\left(\frac{r(x)}{\beta}\right).
    \label{eq:optimal_dist}
\end{equation}
We formulate preference alignment as minimizing $D_{\mathrm{KL}}(p_{\theta} \parallel p^*)$. Utilizing the reparameterization trick \cite{wang2023prolificdreamer, lu2024simplifying}, the gradient to update the generator $G_\theta$ operates entirely in the score domain:
\begin{equation}
    \nabla_\theta \mathcal{L}_{\text{align}} \approx -\mathbb{E}_{t, z} \left[ \Big( \nabla_{x_t} \log p_t^*(x_t) - \nabla_{x_t} \log p_{\theta, t}(x_t) \Big) \frac{dG_\theta(z)}{d\theta} \right].
\label{eq:kl_grad_general}
\end{equation}

\textbf{Approximating the Target Score.} The exact target score $\nabla_{x_t} \log p_t^*(x_t)$ is mathematically intractable. However, we observe that the fake model ($\mathbf{s}_{\mathrm{fake}}$), continuously trained by Eq.~\ref{eq:denoise_loss}, acts as an exceptionally robust denoiser. Based on Tweedie's equivalence \cite{efron2011tweedie, meng2021estimating}, the score can be directly estimated from the denoising prediction. Thus, given a human-preferred sample $x_+$, feeding its noisy variant $F(x_+, t)$ into the fake model yields an implicit projection onto the learned high-reward manifold. This allows us to approximate the intractable target score with
$\mathbf{s}_{\mathrm{fake}}(F(\mathbf{x}_{+},t),t)$, yielding our core DM-Align gradient:
\begin{equation}
\nabla_{\theta} \mathcal{L}_{\text{align}} \approx -\mathbb{E}_{t,z} \left[ \Big( \mathbf{s}_{\mathrm{fake}}\big(F(x_+, t), t \big) - \mathbf{s}_{\mathrm{fake}}\big(F(G_{\theta}(z), t), t \big) \Big) \frac{dG_{\theta}}{d\theta} \right].
\label{eq:align_gradient}
\end{equation}
By operating strictly in the score space, Eq.~\ref{eq:align_gradient} entirely bypasses the need to adapt the diffusion sampling process into a traditional RL framework. Based on this, we design two concrete sample-guided implementations, which are integrated into the main training loop (Algorithm~\ref{alg:training}):

\textbf{DM-PairLoss.} Inspired by DPO, this variant utilizes pre-annotated positive samples (e.g., SFT data) as $x_+$, directly substituting them into Eq.~\ref{eq:align_gradient}. The current generator's output serves as the negative sample, establishing a straightforward contrastive gradient that pulls the generation toward the preferred anchor.

\textbf{DM-GroupLoss.} For reward-model-guided alignment (inspired by GRPO), we dynamically synthesize the alignment target. For a given prompt, we sample a group of $K$ outputs $\{x_1, \dots, x_K\}$ from the generator and compute their rewards and advantages $A_i$. We designate the average of the top-$k$ samples as the preferred anchor $\bar{x}_+$. The gradient adaptively handles the group:
\begin{equation}
\nabla_{\theta}\mathcal{L}_{\text{DM-Group}} \approx -\mathbb{E}_{t, z} \left[ \sum_{i=1}^{K} A_i \Big( \mathbf{s}_{\mathrm{fake}}\big(F(x_i,t),t\big) - \mathbf{s}_{\mathrm{fake}}\big(F(\bar{x}_+,t),t\big) \Big) \frac{dG_{\theta}}{d\theta} \right].
\label{eq:dm_group_gradient}
\end{equation}
This formulation creates a dynamic distribution shift: poor generations ($A_i < 0$) are pulled toward the high-quality group mean $\bar{x}_+$, while superior generations ($A_i > 0$) push the distribution further toward the peak of the preference manifold.

\subsection{Unified Single-Stage Optimization}
\label{sec3.3:unify}
As rigorously outlined in Algorithm~\ref{alg:training}, our framework completely eliminates the traditional barrier between distillation and RL. The generator is optimized via a synergistic combination of both objectives:
\begin{equation}
\label{eq:lambda}
\mathcal{L}_{\text{total}} = \lambda_{\text{DMD}}\mathcal{L}_{\text{DMD}} + \lambda_{\text{align}}\mathcal{L}_{\text{align}}.
\end{equation}
This unification delivers immense computational benefits. 
Specifically, the generator performs the few-step denoising trajectory according to the schedule $\mathcal{T}$ and reuses the generated samples for both objectives. For score estimation, the DMD and DM-Align gradients are evaluated after applying forward noise at an independently sampled level $t \sim \mathcal{U}(0,1)$.
Simultaneously, the generator completes the full few-step denoising process to produce the final output $\mathbf{x}_{\text{final}}$. 
These fully generated samples are then directly utilized as the high-quality negative inputs for DM-PairLoss or the reliable evaluation inputs for DM-GroupLoss.
By seamlessly sharing the latent space and the score estimator ($\mathbf{s}_{\mathrm{fake}}$), DM-Align leverages the acceleration of DMD to generate samples, while DMD benefits from the structural guidance of DM-Align to align with human preferences, achieving mutually enhanced video generation.

\section{Experiments}
\label{sec:Experiments}

\subsection{Experiment Setup}
We primarily employ Wan 2.1-T2V-1.3B \cite{wan2025wan} as our base model, generating 5-second videos at 16 FPS with a resolution of $832 \times 480$. 
We term the pair-based and group-based variants of our method (Section~\ref{sec3:method}) as \textbf{DM-Align (Pair)} and \textbf{DM-Align (Group)}, respectively.

\textbf{Datasets.} 
The choice of distinct datasets is driven by the algorithmic requirements of the two variants. 
For DM-Align (Pair), we adopt the filtered ConsistID dataset \cite{yuan2025identity} (roughly 6K samples). Because DM-Align (Pair) requires high-quality real video pairs to construct contrastive DPO gradients, this human-centric dataset serves as an ideal choice. 
Furthermore, real-world video data inherently contains much more natural, fluid human motions and realistic physical dynamics compared to generated content, providing a superior upper bound for preference alignment. 
Conversely, because DM-Align (Group) utilizes an external reward model to score and guide the optimization process, it naturally supports the use of unannotated textual prompts. Thus, we utilize 20K highly diverse prompts from the expansive VidProM dataset \cite{wang2024vidprom} to broadly cover different generation scenarios and thoroughly explore the reward landscape. 
This strategic dataset selection aligns naturally with the paradigms established in previous RL alignment works \cite{xue2025dancegrpo, liu2025flow}.

\textbf{Baselines and Reward.} 
Baselines include the raw model, standalone DMD2 \cite{yin2024improved}, standalone RL (Flow-DPO \cite{liu2025improving}/DanceGRPO \cite{xue2025dancegrpo}), and a two-stage sequential pipeline (RL + DMD2). 
For DM-Align (Group), we observe that the VideoAlign reward model \cite{liu2025improving} tends to assign erroneously high motion/visual scores to degenerate content (e.g., pure white noise or flickering frames). 
To prevent training collapse for both DanceGRPO and our method, we primarily focus on Textual Alignment (TA) for the VidProM dataset. 
Additionally, to validate the versatility of our framework across diverse reward signals, we employ the image-based HPSv2 model \cite{wu2023human} as an alternative reward in our sensitivity analysis (Section~\ref{sec:sensitivity}). Following standard conventions, the overall video reward is calculated by averaging the per-frame image scores.

\textbf{Training and Evaluation.} 
We conduct all experiments on 32 Nvidia H100 GPUs using the Wan2.1-T2V-1.3B base model. Both the generator and the fake model are optimized with AdamW, using learning rates of $2\times10^{-6}$ and $4\times10^{-7}$, respectively. The fake model $\mathbf{s}_{\mathrm{fake}}$ is updated continuously to provide accurate score estimation, while the generator $G_\theta$ is updated every 5 steps for training stability. We use a per-GPU batch size of 1, start EMA from step 500 with a decay rate of $0.99$, and perform distillation over timesteps $[1000, 750, 500, 250]$ with a teacher CFG scale of $6.0$. For DM-Group, we set the group size to $K=8$ and average the top-4 samples to construct the preference-guided target.
In our combined loss function, $\lambda_{\text{align}}$ and $\lambda_{\text{DMD}}$ are set to 0.5. We provide a comprehensive sensitivity analysis of these weights, including various static ratios and dynamic scheduling, on the CogVideoX model in Section~\ref{sec:sensitivity}.
While a DMD-only warm-up might seem intuitive to prevent early-stage instability, we find it unnecessary. First, even during the initial stages where generations are blurry, the coarse color blocks and basic motion patterns already provide relative advantages to establish a roughly correct direction for preference gradient guidance. 
Second, the DMD gradient inherently dominates this early phase to rapidly establish distribution consistency, natively averting instability without explicit tuning. 
Additionally, we note that standard RL baselines like DanceGRPO and Flow-DPO heavily rely on Exponential Moving Average (EMA) to maintain training stability, without which the visual quality of their generated samples tends to degrade. 
In contrast, our unified framework exhibits robust convergence natively.
While DM-Align and DMD2 converge stably within 500 steps, we train them for 1,000 steps to ensure complete convergence. Conversely, due to the substantially higher per-step time cost of DanceGRPO, we train it for 100 steps.
For evaluation, we curate a test set of 640 samples and adopt VBench \cite{huang2024vbench} for automated metrics alongside the GSB (Good, Same, Bad) protocol for human evaluation.

\subsection{Main Experiments}
The quantitative VBench results are summarized in Table~\ref{tab:main_results}, where NFE denotes the Number of Function Evaluations. 
Overall, standalone DMD2 effectively distills the 100-NFE generation process into a more efficient 4-NFE few-step generation, demonstrating inherent baseline stability. 
Our single-stage approach successfully integrates alignment into this framework without compromising these distillation capabilities.

Specifically, \textbf{DM-Align (Pair)} outperforms all baselines, achieving the highest average score (82.78) with only 4 NFE. 
When analyzing the metrics, both our method and Flow-DPO significantly improve Motion Quality. 
However, the standalone Flow-DPO shows limited gains in Visual Quality (e.g., its Aesthetic Quality drops to 56.32). While the two-stage Flow-DPO + DMD2 pipeline can partially recover this Visual Quality (58.26), it drastically diminishes the Motion Quality benefits (Dynamic Degree drops from 67.13 to 53.43). 
In contrast, our single-stage optimization stably improves both aspects simultaneously, achieving a robust Dynamic Degree of 66.25 and an Aesthetic Quality of 60.89.

For \textbf{DM-Align (Group)}, our method similarly achieves the highest average score (84.40), showing substantial improvements over other baselines. However, similar to the pair variant, the sequential two-stage pipeline suffers from severe performance degradation after distillation. For instance, while DanceGRPO initially achieves a high Dynamic Degree of 75.78, applying DMD2 afterward causes a precipitous drop to 60.85 (a 14.93 reduction). Our approach completely circumvents this post-distillation collapse, pushing the Dynamic Degree to an impressive 80.19.

\begin{table*}[tp]
\centering
\caption{Main results on VBench evaluation. The best results are in \textbf{bold}. The upper section presents results on the ConsistID dataset using DM-Align (Pair), while the lower section shows results on the VidProM dataset using DM-Align (Group).}
\label{tab:main_results}
\resizebox{\textwidth}{!}{%
\begin{tabular}{lcccccccc}
\toprule
\multirow{2}{*}{\makecell[c]{Method}} & \multirow{2}{*}{\makecell[c]{NFE}} & \multirow{2}{*}{\makecell[c]{Average\\Score}} & \multicolumn{2}{c}{Temporal Consistency} & \multicolumn{2}{c}{Motion Quality} & \multicolumn{2}{c}{Visual Quality} \\
\cmidrule(lr){4-5} \cmidrule(lr){6-7} \cmidrule(lr){8-9}
 & & & \makecell{Subject\\Consistency} & \makecell{Background\\Consistency} & \makecell{Motion\\Smoothness} & \makecell{Dynamic\\Degree} & \makecell{Aesthetic\\Quality} & \makecell{Imaging\\Quality} \\
\midrule
\multicolumn{9}{l}{\textit{DM-Align (Pair) on ConsistID Dataset}} \\
\midrule

Wan-T2V-1.3B (raw model)  & 100 & 78.20 & 96.57 & 94.46 & 99.01 & 50.78 & 56.09 & 72.29 \\
DMD2 & 4 & 79.68 & 98.03 & \underline{95.78} & 99.10 & 52.81 & 57.64 & \underline{74.73} \\
Flow-DPO & 100 & \underline{81.54} & 98.10 & 95.56 & \underline{99.25} & \textbf{67.13} & 56.32 & 72.91 \\
Flow-DPO + DMD2 & 4 & 79.89 & \underline{98.14} & \textbf{96.01} & 99.18 & 53.43 & \underline{58.26} & 74.35 \\
\textbf{DM-Align (Pair) (Ours)} & 4 & \textbf{82.78} & \textbf{98.80} & 95.76 & \textbf{99.35} & \underline{66.25} & \textbf{60.89} & \textbf{75.61} \\
\midrule
\multicolumn{9}{l}{\textit{DM-Align (Group) on VidProM Dataset}} \\
\midrule

Wan-T2V-1.3B (raw model) & 100 & 77.85 & 93.93 & 94.63 & 98.63 & 56.40 & 57.33 & 66.20 \\
DMD2 & 4 & 78.88 & 95.53 & 95.22 & \underline{98.60} & 51.64 & \textbf{63.02} & 69.30 \\
DanceGRPO & 100 & \underline{82.76} & 96.04 & 95.86 & 98.33 & \underline{75.78} & 60.44 & 70.15 \\
DanceGRPO + DMD2 & 4 & 80.54 & \underline{96.60} & \underline{95.89} & 98.43 & 60.85 & 60.71 & \underline{70.81} \\
\textbf{DM-Align (Group) (Ours)} & 4 & \textbf{84.40} & \textbf{97.13} & \textbf{96.49} & \textbf{98.99} & \textbf{80.19} & \underline{62.17} & \textbf{71.45} \\
\bottomrule
\end{tabular}
}
\end{table*}

\begin{figure}[tb]
  \centering
  \begin{minipage}[c]{0.56\linewidth}
    \includegraphics[width=\linewidth]{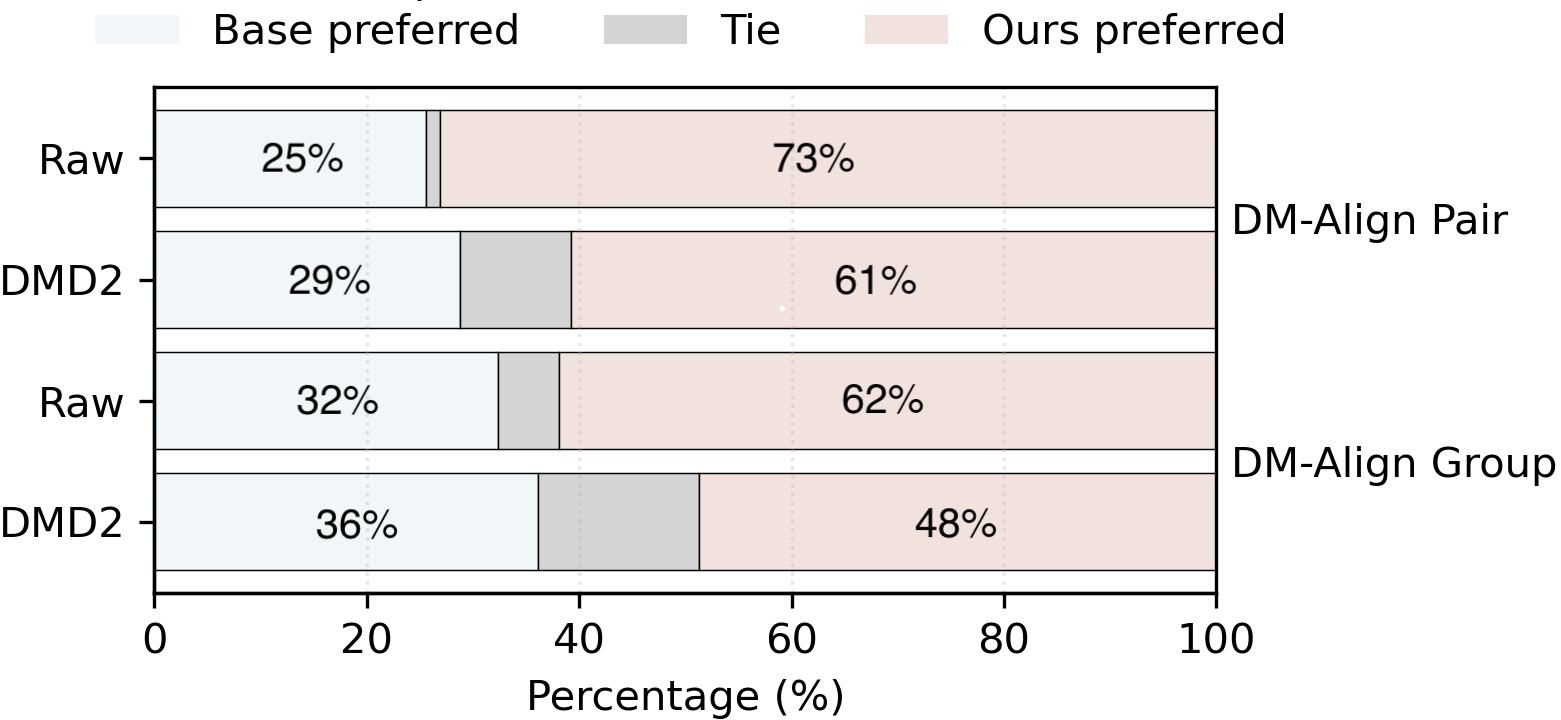}
    \caption{Human evaluation (GSB protocol).}
    \label{fig:gsb_results}
  \end{minipage}
  \hfill
  \begin{minipage}[c]{0.40\linewidth}
    \makeatletter\def\@captype{table}\makeatother 
    \caption{TA scores on VidProM.}
    \label{tab:ta_scores}
    \resizebox{\linewidth}{!}{%
    \begin{tabular}{lc}
    \toprule
    Method & TA Score \\
    \midrule
    Wan-T2V-1.3B (raw) & 0.69 \\
    DMD2 & 0.75 \\
    DanceGRPO & 1.37 \\
    DanceGRPO + DMD2 & 1.40 \\
    \textbf{DM-Align (Group)} & \textbf{1.65} \\
    \bottomrule
    \end{tabular}}
  \end{minipage}
\end{figure}

\begin{figure*}[tp]
    \centering
    \includegraphics[width=\textwidth]{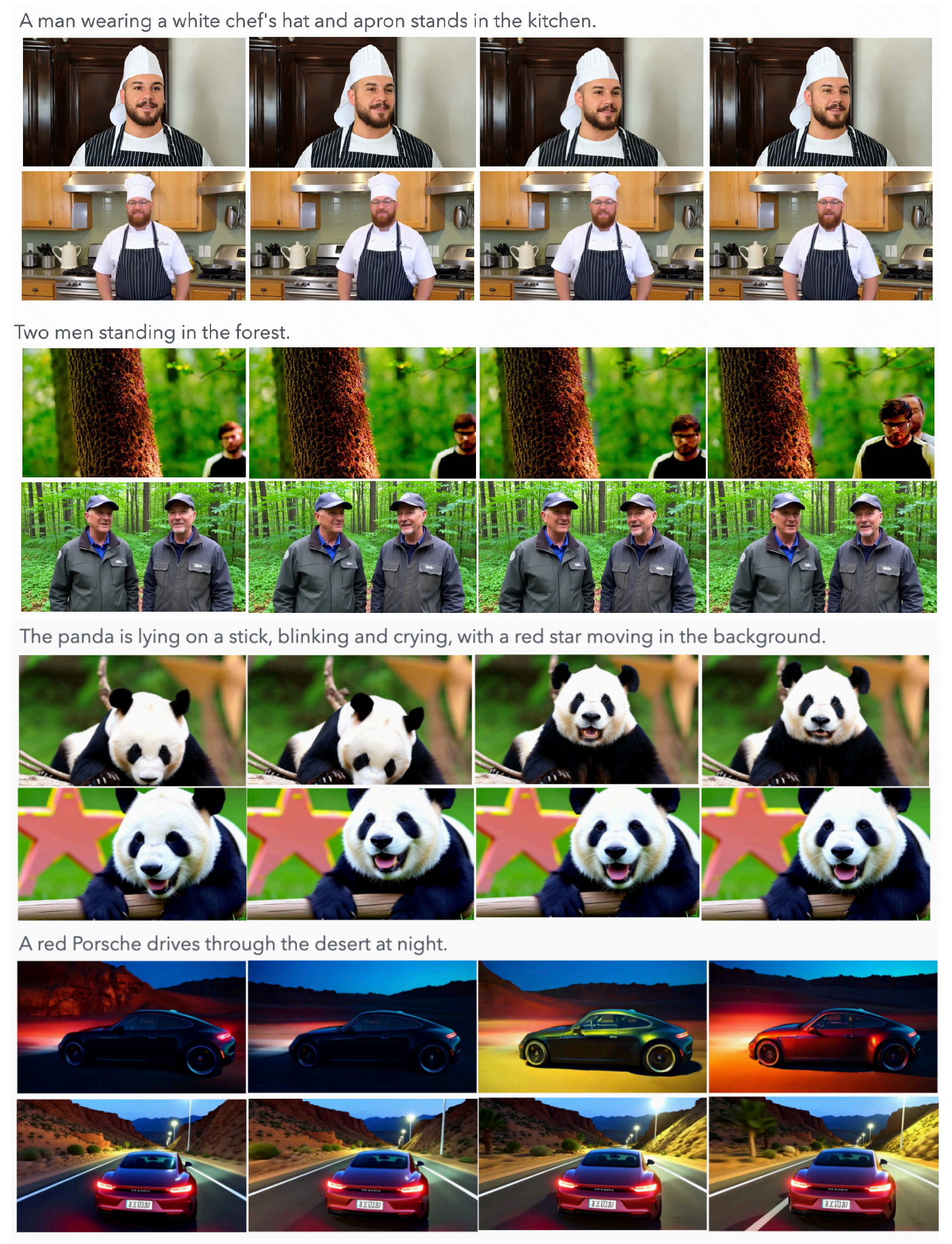} 
    \caption{Qualitative comparison between the raw base model and our DM-Align. For each prompt, the upper row presents the raw base model outputs, and the lower row displays our DM-Align results. 
    The \textit{top two prompts} showcase DM-Align (Pair): whereas the base model produces a frozen, single-person output, our method correctly generates the requested kitchen background and depicts two persons with more natural movements. 
    The \textit{bottom two prompts} showcase DM-Align (Group): our method accurately generates specific textual details, such as a red star and a red Porsche, demonstrating superior prompt consistency.}
    \label{fig:qualitative_main}
\end{figure*}

Human evaluation results (Figure~\ref{fig:gsb_results}) demonstrate that both variants of our method significantly outperform the raw model and DMD2 in overall human preference. 
DM-Align (Pair) shows particularly pronounced improvements (+48\% and +32\% net preference) as it leverages real human preference data, providing highly accurate gradient directions for alignment. 
Although DM-Align (Group) relies on multiple sampling and reward model estimations for preference-guided gradients, it still delivers substantial performance gains (+30\% and +12\% net preference). 
Furthermore, regarding Textual Alignment (TA)—the primary reward metric focused on during DM-Align (Group) training—Table~\ref{tab:ta_scores} confirms that our framework provides superior guidance, drastically outperforming the next best baseline (1.65 vs. 1.40). 
As illustrated in Figure~\ref{fig:qualitative_main}, our approach exhibits clear visual improvements in both prompt consistency and motion smoothness compared to the base model outputs.

\subsection{Efficiency, Scalability, and Generality}
\label{sec:sensitivity}
Beyond the primary VBench metrics, we extend our evaluation to validate the core architectural advantages of DM-Align.

\textbf{Training Efficiency.} As shown in Table~\ref{tab:efficiency}, traditional RL methods like DanceGRPO incur prohibitive multi-step sampling costs, requiring over 1500 NFE and 400 seconds per optimization step. In stark contrast, our single-stage DM-Align completely eliminates multi-step sampling bottlenecks, reducing the training time by nearly $30\times$ while natively aligning the model.

\textbf{Scalability to MoE Architectures.} To validate scalability, we conducted experiments on the latest large-scale Mixture-of-Experts (MoE) architecture, \textbf{Wan2.2-T2V-A14B} \cite{wan2025wan}. The Wan2.2 architecture consists of two distinct 14B models for high-noise and low-noise regimes. Empirically, we observe that applying DM-Align solely to the high-noise expert is sufficient to achieve substantial gains (Table~\ref{tab:moe_results}). Since high-level semantic consistency is predominantly determined in the high-noise regime, aligning only this expert is both remarkably effective and computationally efficient.

\begin{table}[tb]
\centering
\begin{minipage}[t]{0.52\linewidth}
    \centering
    \caption{Training efficiency comparison. `Time (s)' denotes the time per optimization step in seconds.}
    \label{tab:efficiency}
    \begin{tabular}{lcc}
    \toprule
    Method & NFE & Time (s) \\
    \midrule
    DMD2 &  46  & 11 \\
    DanceGRPO & 1536 & 423 \\
    \textbf{DM-Align (Ours)} & \textbf{51} & \textbf{15} \\
    \bottomrule
    \end{tabular}
\end{minipage}
\hfill
\begin{minipage}[t]{0.40\linewidth}
    \centering
    \caption{VBench Scores on the latest MoE model (Wan2.2-A14B).}
    \label{tab:moe_results}
    \begin{tabular}{lcc}
    \toprule
    Method  & NFE & Score \\
    \midrule
    Raw Base  & 100 & 80.00 \\
    DMD   & 4 & 78.35 \\
    \textbf{DM-Align}  & 4 & \textbf{83.45} \\
    \bottomrule
    \end{tabular}
\end{minipage}
\end{table}

\begin{table}[tb]
\centering
\caption{Generality \& Sensitivity analysis ($\lambda_{\text{DMD}} : \lambda_{\text{align}}$) across different backbones and reward models. ``Varying'' denotes a linear schedule where the weight ratio
$\lambda_{\mathrm{DMD}}:\lambda_{\mathrm{align}}$
is linearly scheduled from $1:0$ to $1:3$ over training.
}
\label{tab:ablation}
\resizebox{0.85\linewidth}{!}{
\begin{tabular}{llccccc}
\toprule
Base Model & Reward & 1:0 (DMD) & 3:1 & \textbf{1:1 (Ours)} & 1:3 & Varying \\
\midrule
\multirow{2}{*}{Wan-1.3B}  & TA ($\uparrow$) & 0.75 & 1.34 & \textbf{1.65} & \underline{1.56} & 1.45 \\ 
& HPSv2 ($\uparrow$) & 20.50 & 24.01 & \textbf{28.89} & 12.37 & \underline{27.56} \\ 
\midrule
\multirow{2}{*}{CogVideoX}  & TA ($\uparrow$) & 0.22 & \textbf{0.58} &  \underline{0.55} & 0.08 & 0.49 \\ 
& HPSv2 ($\uparrow$) & 18.71 & 20.92 & \textbf{25.01} & 13.78 & \underline{22.11} \\ 
\bottomrule
\end{tabular}}
\end{table}

\textbf{Generality and Sensitivity.} We extended DM-Align to \textbf{CogVideoX-2B} \cite{yang2024cogvideox} and evaluated it using an alternative reward model (\textbf{HPSv2}). 
As presented in Table~\ref{tab:ablation}, DM-Align successfully boosts performance over the standalone DMD baseline on both models and metrics without specific hyper-parameter tuning. 
Furthermore, we conducted a sensitivity analysis on the weight ratio $\lambda_{\text{DMD}} : \lambda_{\text{align}}$, keeping their sum constrained to 1. 
Alongside various static ratios, we also test a dynamic “Varying” strategy, where the ratio $\lambda_{\mathrm{DMD}}:\lambda_{\mathrm{align}}$ is linearly scheduled from 1:0 to 1:3 over the training steps, while normalizing the two weights to maintain $\lambda_{\mathrm{DMD}}+\lambda_{\mathrm{align}}=1$. This schedule starts from pure distillation to establish structural consistency and gradually increases the alignment strength to refine human preferences.
The results reveal that performance remains highly robust across a wide range of configurations. Severe instability or degradation only occurs when the DMD weight is excessively low (e.g., 1:3), where the insufficient distillation signal fails to sustain the base few-step generation capability.

\begin{figure}[tb]
  \centering
  \begin{minipage}[t]{0.40\linewidth}
    \centering
    \includegraphics[width=\linewidth]{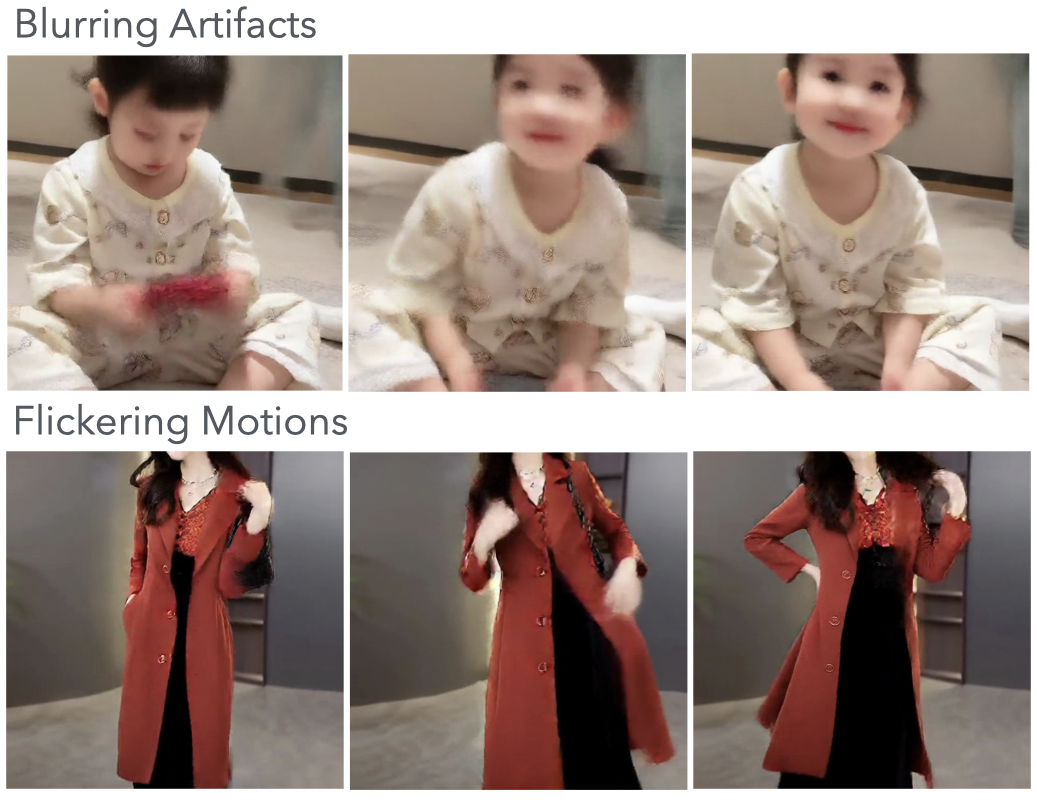} 
    \caption{Visual examples of catastrophic model collapse when attempting to fine-tune a distilled model (the sequential ``RL-after-Distillation'' pipeline).}
    \label{fig:distill_after}
  \end{minipage}
  \hfill
  \begin{minipage}[t]{0.55\linewidth}
    \centering
    \includegraphics[width=\linewidth]{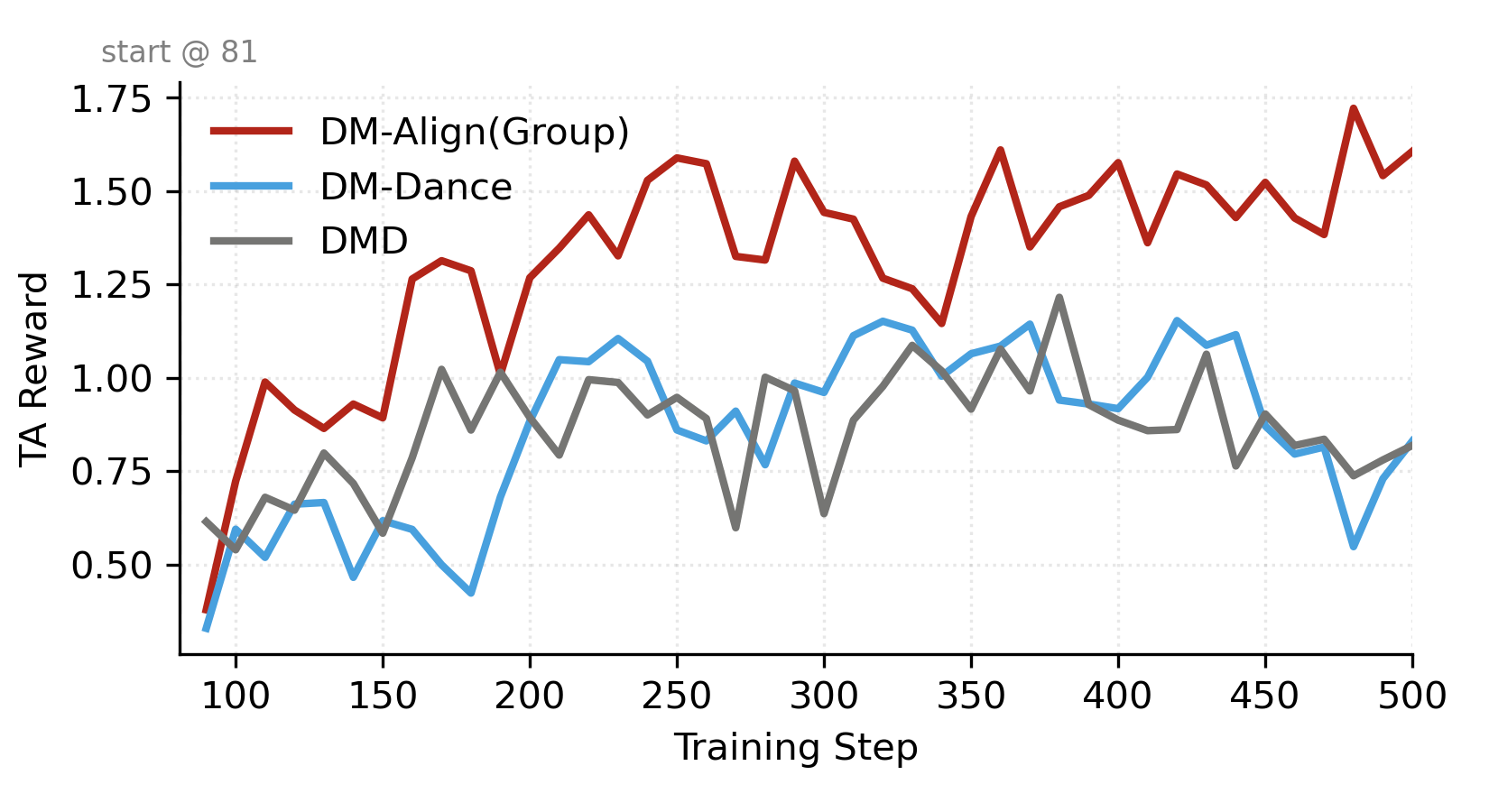}
    \caption{Reward progression during training. Unlike direct RL loss integration, DM-Align provides superior and stable reward guidance.}
    \label{fig:reward}
  \end{minipage}
\end{figure}

\subsection{Analysis of Alternative Approaches}

\textbf{Why Not Sequential Pipelines?} 
An intuitive approach is the two-stage pipeline: either ``Distillation-after-RL'' or ``RL-after-Distillation''. 
As demonstrated in Table~\ref{tab:main_results}, optimizing alignment and distillation in two isolated stages often results in objective misalignment. Specifically, applying distillation to a previously RL-aligned model tends to degrade the very metrics (e.g., motion quality and dynamic degree) that the initial RL stage sought to improve. 
Conversely, performing RL training \textit{after} the distillation process to leverage fast sampling is equally catastrophic. 
As shown in Figure~\ref{fig:distill_after}, fine-tuning a distilled model leads to rapid degradation, manifesting as noticeable blurring artifacts and flickering motions, eventually progressing to complete model collapse. 
We hypothesize that a distilled model's parameters are hypersensitive; even minor gradient adjustments from sparse RL signals shatter its few-step generation capability. 
This reinforces the necessity of our unified, single-stage distribution matching framework.

\textbf{Why Not Direct RL Loss Integration?} 
Another straightforward alternative to our method would be directly incorporating off-the-shelf RL losses alongside the DMD optimization process. 
To investigate this, we developed a variant that combines the DanceGRPO loss with the standard DMD loss. 
As illustrated in Figure~\ref{fig:reward}, this direct RL loss variant performs nearly identically to the standalone DMD, struggling to provide meaningful reward progression.
We attribute this to the overly conservative gradient updates characteristic of standard RL losses designed for visual generation. Because applying standard RL directly to continuous generative models is notoriously unstable, their loss formulations are heavily constrained. 
In contrast, our DM-Align operates natively in the distribution matching space, leveraging the fake model's unified denoising vector field to provide dense, stable, and highly compatible guidance.

\textbf{Does the Gain Come from the Dataset?}
A potential concern is that the improvement of DM-Align may primarily stem from the high-quality ConsistID data rather than the proposed unified formulation. To disentangle this dataset factor, we conduct a prompt-only cross-evaluation on ConsistID, where DM-Group uses only the textual prompts without accessing the paired real videos. As shown in Table~\ref{tab:cross_eval}, DM-Group still substantially outperforms DMD2, the sequential RL baseline DanceGRPO, and the concurrent joint-optimization method DMDR. This result suggests that the performance gain is not merely a byproduct of high-quality paired data, but arises from formulating preference alignment as a native score-space distribution-matching gradient compatible with DMD.

\begin{table}[t]
\centering
\caption{Cross-evaluation on ConsistID using prompts only. DM-Group uses only textual prompts without paired real videos.}
\label{tab:cross_eval}
{
\begin{tabular}{lccc}
\toprule
\textbf{Method} & \textbf{Dynamic} & \textbf{Aesthetic} & \textbf{Temporal} \\
\midrule
DMD2 & 74.90 & 64.19 & 95.51 \\
DanceGRPO & 78.42 & 66.85 & 96.12 \\
DMDR~\cite{jiang2025distributionmatchingdistillationmeets} & 77.15 & 67.40 & 95.88 \\
\midrule
\textbf{DM-Group} & \textbf{81.35} & \textbf{68.12} & \textbf{97.15} \\
\bottomrule
\end{tabular}
}
\end{table}

\subsection{Limitations and Discussion}
\label{sec:discussion}
While our framework successfully bridges distillation and preference alignment, several limitations remain. 
First, although we provide a theoretical derivation for the alignment gradient, the joint optimization dynamics of the weighted loss summation lack strict theoretical convergence guarantees. 
Second, our implementation is built upon the foundational DMD framework. This restricts the distillation mechanism to DMD-based techniques, excluding other recent GAN-based alternatives \cite{mao2025osv, lin2025diffusion, cheng2025pose} that have shown promising distillation performance.

However, our primary algorithm design intentionally prioritizes simplicity and foundational validation. Recently, advanced DMD-based enhancements have emerged \cite{shao2025magicdistillation, luo2025learning, fan2025phaseddmdfewstepdistribution}. 
Because our DM-Align operates orthogonally as a native, score-based gradient provider, it can be seamlessly integrated with these advancements to yield further improvements. We leave these exciting integrations for future work. 
More broadly, we hope this work invites a higher-level perspective on distillation and alignment—transitioning them from traditionally separated stages into a unified paradigm.

\section{Conclusion}
\label{sec:conclusion}
In this paper, we introduced DM-Align, a novel framework that unifies video generation model distillation and human preference alignment into a single cohesive stage through distribution matching. 
By operating entirely in the score domain, our approach eliminates the need to adapt the diffusion sampling process into an MDP formulation, simultaneously bypassing time-intensive multi-step generation bottlenecks that plague conventional diffusion RL methods. 
Extensive evaluations across diverse base models and reward signals demonstrate that our method preserves the efficiency of few-step distillation while securing robust alignment with human preferences and exhibiting strong generalization capabilities. 
Both automated metrics and human evaluations confirm that DM-Align significantly outperforms standalone baselines and conventional two-stage pipelines, paving the way for more efficient and unified generative model optimization.

\section*{Acknowledgment}

This work was supported in part by the Shenzhen Science and Technology Program\textbf{}
(Grant No.~RCJC20210706091946001), and in part by the Shenzhen Science and
Technology Program (Grant No.~ZDCY20250901104207008).



%
%
\bibliographystyle{splncs04}
\bibliography{main}
\end{document}